\documentclass{article}
\usepackage{amsmath,amsfonts,amsthm,algorithm,algorithmic,graphicx,subfigure}

\usepackage[sort&compress]{natbib}

\usepackage{hyperref}

\bibpunct{(}{)}{;}{a}{,}{,}

\newcommand{\PPP}{\mathcal{P}}				
\newcommand{\R}{\mathbb{R}}				
\newcommand{\BP}{\begin{proof}}			    
\newcommand{\EP}{\end{proof}}			

\makeatletter
\floatstyle{\ALG@floatstyle}
\newfloat{protocol}{htbp}{loa}
\floatname{protocol}{Protocol}
\makeatother

\newtheorem{theorem}{Theorem}
\newtheorem*{theorem*}{Theorem}
\newtheorem{proposition}{Proposition}
\newtheorem{lemma}{Lemma}
\newtheorem{corollary}{Corollary}

\theoremstyle{definition}
\newtheorem{definition}{Definition}

\begin{document}

\title{Linear Probability Forecasting}
\date{}

\author{Fedor Zhdanov and Yuri Kalnishkan\\
Computer Learning Research Centre,\\
Department of Computer Science,\\
Royal Holloway University of London,\\
Egham, Surrey, TW20 0EX, UK\\
\{fedor,yura\}@cs.rhul.ac.uk\\
}

\maketitle
\thispagestyle{empty}

\begin{abstract}
Multi-class classification is one of the most important tasks in machine learning. In this paper we consider two online multi-class classification problems: classification by a linear model and by a kernelized model. The quality of predictions is measured by the Brier loss function. We suggest two computationally efficient algorithms to work with these problems and prove theoretical guarantees on their losses. We kernelize one of the algorithms and prove theoretical guarantees on its loss. We perform experiments and compare our algorithms with logistic regression.
\end{abstract}

\section{Introduction}
Online prediction is a wide area of machine learning (see \citealp{CesaBianchi2006}). Its algorithms can be applied to different data mining problems (see for example \citealp{Freund1997}). Online prediction provides efficient algorithms which adapt to a predicted process ``on fly''. In online regression framework we assume the existence of some input at each step and try to predict an outcome on this input. This process is repeated step by step. We consider multi-dimensional Brier game where outcomes and predictions come from a simplex and can be thought of as probability distributions on the vertices of the simplex. If the outcomes are identified with vertices of the simplex this problem can be thought of as the multi-class classification problem of the given input.

In the simple case the dependence between the input and its outcome is assumed to be linear; linear regression minimising the expected loss is studied in statistics. As opposite to the traditional statistical setting, the learner in online prediction does not make any statistical assumptions about the data generating process. Its goal is to predict as well as the best linear function on input. Instead of looking for the best linear function, our learner considers all linear functions and makes his prediction by mixing them in a certain way at each prediction step. We prove theoretical bounds on the cumulative loss of the learner in comparison with the cumulative loss of the best linear function (we say the learner competes with these functions). We consider the square loss: mean square error is one of the benchmark measures for classification algorithms (see \citealp{Brier1950}).

We use Vovk's Aggregating Algorithm (a generalization of the Bayesian mixture) to mix functions (as in Aggregating Algorithm Regression, AAR: see \citealp{VovkCOS}). This method has previously been applied to the case when possible outcomes lie in a segment of the real line, and so the prediction was one-dimensional. We develop two algorithms to solve the problem of multi-dimensional prediction. The first algorithm applies a variant of AAR to predict each coordinate of the outcome separately, and then combines these predictions in a certain way to get probability prediction. The other algorithm is designed to give probability predictions directly; these are first computationally efficient online regression algorithm designed to solve linear and non-linear multi-class classification problems. We derive theoretical bounds on the losses of both algorithms. We come to an unexpected conclusion that the component-wise algorithm is better than the second one asymptotically, but worse in the beginning of the prediction process. Their performance on benchmark data sets is very similar.

One component of the prediction of the second algorithm has the meaning of a remainder. In practice this situation is quite common. For example, in a football match either one team wins or the other, and the remainder is a draw (see \citet{VovkPEABG} for online prediction experiments in football). When we analyse a precious metal alloy we may look for a description of the following kind: the alloy has 40\% of gold, 35\% of silver, and some addition (e.g., copper and palladium). It is common for financial applications to predict the direction of the price: the price can go up, down, or stay close to the current value. We perform classification experiments with linear algorithms and compare them with logistic regression.

A description of the framework can be found in Section~\ref{sec:frame}, description of the algorithms can be found in Section~\ref{sec:linearalg}, and derivation of the theoretical bounds can be found in Section~\ref{sec:linearbound}.

We look for a way to extend the class of experts using the kernel trick. We kernelize the second algorithm and prove a theoretical bound on its loss. The cumulative loss of the kernelized algorithm is compared with the cumulative loss of any finite set of functions from the RKHS given by a kernel parameter it uses. Kernelization process is described in Section \ref{sec:kernel}. Our experiments are shown in Section~\ref{sec:experiments}. Section~\ref{sec:conclusion} makes the conclusions and shows some possibilities for prospective work.

\section{Framework}\label{sec:frame}
A game of prediction contains three components: a space $\Omega$ of outcomes, a decision space $\Gamma$, and a loss function $\lambda:\Omega\times\Gamma\to \R$. We are interested in the generalisation of the Brier game from \cite{Brier1950} where the space of outcomes $\Omega = \PPP(\Sigma)$ is the set of all probability measures on a finite set $\Sigma$ with $d$ elements, $\Gamma:=\{(\gamma_1,\ldots,\gamma_d): \sum_{i=1}^d \gamma_i = 1, \gamma_i \in \R\}$ is a hyperplane in $d$-dimensional space containing all the outcomes, and for any $y\in \Omega$ we define the loss
\begin{equation*}
  \lambda(y,\gamma)
  =
  \sum_{\sigma\in\Sigma}
  \left(
    \gamma\{\sigma\} - y\{\sigma\}
  \right)^2.
\end{equation*}
For example,
if $\Omega=\{1,2,3\}$, $\omega=1$,
$\gamma\{1\}=1/2$, $\gamma\{2\}=1/4$, and $\gamma\{3\}=1/4$,
$\lambda(\omega,\gamma)=(1/2-1)^2+(1/4-0)^2+(1/4-0)^2=3/8$.
Brier loss is one of the most important loss functions used to assess the quality of classification algorithms. The game of prediction is being played repeatedly by a learner receiving some input vectors $x_t \in \mathbf{X} \subseteq \mathbb{R}^n$, and follows prediction protocol~\ref{prot:PEA}.

\begin{protocol}[h]
  \caption{Protocol of forecasting game}
  \label{prot:PEA}
  \begin{algorithmic}
    \STATE $L_0:=0$.
    \FOR{$t=1,2,\dots$}
      \STATE Reality announces a signal $x_t \in \mathbf{X} \subseteq \R^n$.
      \STATE Learner announces $\gamma_t \in \Gamma \subseteq \R^d$.
      \STATE Reality announces $y_t\in\Omega \subseteq \R^d$.
      \STATE $L_t:=L_{t-1}+\lambda(y_t,\gamma_t)$.
    \ENDFOR
  \end{algorithmic}
\end{protocol}

We find an algorithm which is capable of competing with all linear functions (we call them experts) $\xi_t=(\xi^1_t,\dots,\xi^d_t)'$ on $x$:
\begin{align}\label{eq:experts1}
\xi^1_t &= 1/d + \alpha_1 ' x_t\notag\\
&\ldots \notag\\
\xi^{d-1}_t &= 1/d + \alpha_{d-1} ' x_t\\
\xi^d_t &= 1 - \xi^1 - \dots - \xi^{d-1} = 1/d - \left(\sum_{i=1}^{d-1}\alpha_i\right) ' x_t, \notag
\end{align}
where $\alpha_i=(\alpha_i^1,\dots,\alpha_i^n)',\enspace i=1,\ldots,d-1$. In the model~\eqref{eq:experts1} the prediction for the last component of an outcome is calculated from the predictions for other components. Denote $\alpha = (\alpha_1',\dots,\alpha_{d-1}')' \in \Theta = \mathbb{R}^{n(d-1)}$. Then any expert can be presented as $\xi_t=\xi_t(\alpha)$. Let also $L_T(\alpha)=\sum_{t=1}^T \lambda(y_t,\xi_t(\alpha))$ be the cumulative loss of an expert $\alpha$ over $T$ trials.

\section{Derivation of the algorithms}\label{sec:linearalg}
In this section we describe how we apply the Aggregating Algorithm (AA) proposed in \cite{VovkAS} to mix experts and make predictions. The algorithm keeps weights $P_{t-1}(d\alpha)$ for the experts at each prediction step $t$, and updates them by the exponential weighting scheme after the actual outcomes is announced:
\begin{equation}\label{eq:wupdate}
P_t(d\alpha)=\beta^{\lambda(y_t,\xi_t(\alpha))}P_{t-1}(d\alpha), \quad \beta \in (0,1).
\end{equation}
Here $\beta = e^{-\eta}$, where $\eta \in (0,\infty)$ is a learning rate parameter. This weight update ensures that the experts which predict badly at the step $t$ receive less weight. The weights are then normalized $P_t^*(d\alpha) = \frac{P_t(d\alpha)}{P_t(\Theta)}$.

The prediction of the algorithm is a combination of the experts' predictions. It is suggested in \citet{Kivinen1999} that the prediction is simply the weighted average of the experts' predictions with weights $P_t(d\alpha)$. The Aggregating Algorithm uses more sophisticated prediction scheme, and sometimes achieves better theoretical performance. It first defines a \emph{generalised prediction} at any step $t$ as a function $g_t: \Omega \to \mathbb{R}$ such that
\begin{equation}\label{eq:genpred}
g_t(y) = \log_\beta \int_{\Theta} \beta^{\lambda(y,\xi_t(\alpha))}P^*_{t-1}(d\alpha)
\end{equation}
for all $y \in \Omega$. It is a weighted average (in a general sense) of the experts' losses for each possible outcome. It then predicts any $\gamma_t$ such that
\begin{equation}\label{eq:lAA}
    \lambda(y,\gamma_t) \le g_t(y)
\end{equation}
for all possible $y \in \Omega$. If such prediction can be found for any weights distribution on experts the game is called \emph{perfectly mixable}. Perfectly mixable games and other types of games are analyzed in \citet{VovkGofPEA}. It is also shown there that for countable (and thus finite) number of experts the AA achieves the best possible theoretical guarantees.

\subsection{Proof of mixability}\label{ssec:mixabil}
In this section we prove that our game is perfectly mixable and show a function that can be used to give predictions satisfying~\eqref{eq:lAA}.

It is shown in Theorem 1 \cite{VovkPEABG} that the Brier game with finite number of outcomes is perfectly mixable iff $\eta \in (0,1]$.
The two authors of that paper consider the outcome space of $d$ probability measures concentrated in points of $\Sigma$. We denote this space by $\mathcal{R}(\Sigma)$. They consider experts giving predictions from all probability measures $\PPP(\Sigma)$. We need to prove that the inequality \eqref{eq:lAA} holds for our experts \eqref{eq:experts1} (who can give predictions outside of the probability simplex) and our outcome space $\Omega$ (the whole probability simplex, not only its vertices). Lemma \ref{lem:mixab} describes the first part, but first we need to state an additional statement. The following lemma shows that any vector from $\R^d$ can be projected into simplex without increasing the Brier loss.
\begin{lemma}\label{lem:projection}
  For any $\xi = (\xi_1,\ldots,\xi_d) \in \R^d$ there exists $\theta = (\theta_1,\ldots,\theta_d) \in \PPP(\Sigma)$ such that for any $y \in \Omega$ we have $\lambda(y,\theta) \le \lambda(y,\xi)$.
\end{lemma}
\BP
  The Brier loss of a prediction $\gamma$ is a square Euclidean distance between $\gamma$ and the actual outcome $y$ in a $d$-dimensional space. The proof follows from the fact that $\Omega$ is a convex and closed set in $\R^d$.
\EP

\begin{lemma}\label{lem:mixab}
  Let $P(d\alpha)$ be any probability distribution on $\Theta$. Then for any $\eta \in (0,1]$ there exists $\gamma \in \Gamma$ such that for any $y \in \mathcal{R}(\Sigma)$ we have
  \begin{equation*}
    \lambda(y,\gamma) \le \log_\beta \int_{\Theta} \beta^{\lambda(y,\xi(\alpha))}P(d\alpha).
  \end{equation*}
\end{lemma}
\BP
  By Lemma \ref{lem:projection} for any $\xi(\alpha)$ we can find $\theta(\alpha) \in \PPP(\Sigma)$ such that the loss of experts decreases: $\lambda(y,\theta(\alpha)) \le \lambda(y,\xi(\alpha))$ for any $y \in \mathcal{R}(\Sigma)$. Thus we have
  \begin{equation*}
  \log_\beta \int_{\Theta} \beta^{\lambda(y,\theta(\alpha))}P(d\alpha) \le \log_\beta \int_{\Theta} \beta^{\lambda(y,\xi(\alpha))}P(d\alpha)
  \end{equation*}
  for any $y \in \mathcal{R}(\Sigma)$. We can take the same prediction $\gamma \in \Gamma$ that satisfies the necessary inequality with $\theta$ instead of $\xi$. By Theorem 1 in \cite{VovkPEABG} such prediction exists for any $\eta \in (0,1]$ ($\beta \in [e^{-1},1)$).
\EP
A way to convert the generalised prediction into the prediction of AA is called a substitution function. We prove that we can use the same substitution function and the same learning rate parameter $\eta$ as for the case of finite number of possible outcomes. Such a function is proposed in \cite{VovkPEABG}. This is an extension of Lemma 4.1 from \cite{Haussler1998}.
\begin{lemma}\label{lem:contout}
  Let $P(d\alpha)$ be a probability distribution on $\Theta$ and put
  \begin{equation*}
    f(y) = \log_\beta \int_{\Theta} \beta^{ \lambda(y,\xi(\alpha))} P(d\alpha)
  \end{equation*}
  for every $y \in \Omega$. Then if $\gamma$ is such a prediction that $\lambda(z,\gamma) \le f(z)$ for any $z \in \mathcal{R}(\Sigma)$ then $\lambda(y,\gamma) \le f(y)$ for any $y \in \Omega$.
\end{lemma}
\BP
For the typographical reasons we will write $\xi$ instead of $\xi(\alpha)$. It is easy to ensure that
$
\lambda(y,\gamma) - \lambda(y,\xi) = \sum_{\sigma \in \Sigma} y\{\sigma\}[\lambda(z_\sigma,\gamma) - \lambda(z_\sigma,\xi)]
$
for $z_\sigma\{\rho\}=0$ if $\sigma \ne \rho$ and $z_\sigma\{\rho\}=1$ if $\sigma = \rho$. We also have that $\lambda(y,\gamma) - f(y) \le 0$ is equivalent to $\int_\Theta \beta^{\lambda(y,\xi)-\lambda(y,\gamma)}P(d\alpha) \le 1$. Thus due to the convexity of the exponent function
$
\int_\Gamma \beta^{\sum_{\sigma \in \Sigma} y\{\sigma\}[\lambda(z_\sigma,\xi) - \lambda(z_\sigma,\gamma)]}P(d\alpha)
\le \sum_{\sigma \in \Sigma} y\{\sigma\} = 1.
$
\EP
Let us denote the $i$-th possible outcome from $\mathcal{R}(\Sigma)$ by $y\{i\}, i=1,\ldots,d$. We use the substitution function defined by the following proposition:
\begin{proposition}\label{prop:main}
  Let $r_i = g(y\{i\})$, and $x^+ = \max(x,0)$. Define $s\in \mathbb{R}$ by the requirement
  \begin{equation*}
    \sum_{i=1}^d
    (s-r_i)^+
    =
    2.
  \end{equation*}
  If the prediction of the Aggregating Algorithm is given by
  \begin{equation*}
    \gamma^i = \frac{(s-r_i)^+}{2},\enspace i=1,\ldots,d
  \end{equation*}
  then \eqref{eq:lAA} holds.
\end{proposition}
This function allows us to avoid weights normalization in calculating the generalized prediction at each step (avoid $^*$ in the weights distribution), which would be computationally inefficient. Suppose we can get only $r=g_t(y) + C$ instead of $g_t(y)$, where $C$ is the same for all $y$. Then predictions $\gamma_t$ defined by the substitution function from Proposition~\ref{prop:main} will be the same as if we calculated the generalized prediction with weights normalization.

\subsection{Algorithm for multidimensional outcomes}\label{ssec:multidim}
We set the prior weights distribution $P_0$ over the set $\Theta=\mathbb{R}^{n(d-1)}$ of experts $\alpha$ to have the Gaussian density with a parameter $a > 0$:
\begin{equation*}
(a\eta/\pi)^{n(d-1)/2}e^{-a\eta\|\alpha\|^2}d\alpha.
\end{equation*}
Instead of taking the integral in~\eqref{eq:genpred} we get a shifted generalised prediction $r$ by calculating $r_i = g_T(y\{i\}) - g_T(y\{d\})$ (we omit the index $T$ in $r$ for brevity). Each component of $r = (r_1,\ldots,r_d)$ corresponds to one of the possible outcomes, so $r_d = 0$. Other components, $i=1,\ldots,d-1$:
\begin{equation*}
   r_i = \log_\beta{\frac{ \beta^{g_T(y\{i\}) + \sum_{t=1}^{T-1}g_t(y_t)} }{ \beta^{g_T(y\{d\})+ \sum_{t=1}^{T-1}g_t(y_t)}}}
       =\log_\beta\frac{\int_\Theta e^{-\eta Q(\alpha,y\{i\})}d\alpha}{\int_\Theta e^{-\eta Q(\alpha,y\{d\})}d\alpha}
\end{equation*}
where by $Q(\alpha,y)$ we denote the quadratic form:
\begin{equation*}
Q(\alpha,y) =\sum_{t=1}^{T}\sum_{i=1}^{d}((y_t^i-\xi^i(x_t))^2.
\end{equation*}
Here $y_t = (y_{t}^1,\ldots,y_{t}^d)$ are the outcomes on the steps before $T$ and $y_T = (y_T^1,\ldots,y_T^d)$ is a \emph{possible} outcome on the step $T$.

Let $C = \sum_{t=1}^T x_t x_t '$ be $n \times n$ matrix. The quadratic form $Q$ can be divided into a quadratic part, a linear part, and a remainder: $Q=Q_1 + Q_2 + Q_3$. Here
\begin{equation*}
Q_1(\alpha,y) = \alpha' A \alpha
\end{equation*}
is a quadratic part of $Q(\alpha,y)$. Here $A$ is a square matrix with $n(d-1)$ rows (see the expression for $A$ in the algorithm below). The linear part is equal to
\begin{equation*}
    Q_2(\alpha,y) = h'\alpha - 2\sum_{i=1}^{d-1} (y_T^i - y_T^d)\alpha_i ' x_T,
\end{equation*}
where $h_i = -2\sum_{t=1}^{T-1}(y_t^i - y_t^d)x_t, i=1,\ldots,d-1$ make up a big vector $h=(h_1',\dots,h_{d-1}')'$. The remainder is equal to
\begin{equation*}
    Q_3(\alpha,y)
    = \sum_{t=1}^{T-1} \sum_{i=1}^{d} (y_t^i - 1/d)^2 + \sum_{i=1}^{d} (y_T^i - 1/d)^2.
\end{equation*}

Ratio for $r_i$ can be calculated using the following lemmas. The integral evaluates as follows:
\begin{lemma}\label{lem1}
  Let $Q(\alpha)=\alpha'A\alpha + b'\alpha + c$, where $\alpha,b \in \mathbb{R}^n$, $c$ is a scalar and $A$ is a symmetric positive definite $n \times n$ matrix. Then
  \begin{equation*}
    \int_{\mathbb{R}^n} e^{-Q(\alpha)} d\alpha = e^{-Q_0} \frac{\pi^{n/2}}{\sqrt{\det A}},
  \end{equation*}
  where $Q_0 = \min_{\alpha \in \R^n} Q(\alpha)$.
\end{lemma}
The proof of this lemma can be found in \citet[Theorem 15.12.1]{Harville1997}.
Following this lemma, we can rewrite $r_i$ as $r_i = F(A, b_i, z_i),\enspace i=1,\ldots,d-1,$ where
\begin{equation*}
    F(A, b_i, z_i)=\min_{\alpha \in \Theta}Q(\alpha,y^i)-\min_{\alpha \in \Theta}Q(\alpha,y^d).
\end{equation*}
Variables $b_i,z_i$ and the precise formula for $F$ are defined by the following lemma
\begin{lemma}\label{lem2}
  Let
  \begin{equation*}
    F(A,b,z) = \min_{\alpha \in \R^n}(\alpha' A \alpha + b'\alpha + z'\alpha)
    - \min_{\alpha \in \R^n}(\alpha' A \alpha + b'\alpha - z'\alpha),
  \end{equation*}
  where $b,z \in \R^n$ and $A$ is a symmetric positive definite $n \times n$ matrix. Then $F(A,b,z) = -b'A^{-1}z$.
\end{lemma}
\BP
This lemma is proven by taking the derivative of the quadratic forms in $F$ by $\alpha$ and calculating the minimum: $\min_{\alpha \in \mathbb{R}^n}(\alpha' A \alpha + c'\alpha) = -\frac{(A^{-1}c)'}{4}c$ for any $c \in \R^n$ \citep[see][Theorem 19.1.1]{Harville1997}.
\EP

We can see that $b_i = h + (x_T',\dots,x_T',\mathbf{0},x_T',\dots,x_T')' \in \R^{n(d-1)}$, where $\mathbf{0}$ is a zero-vector from $\R^n$. We also have $z_i = (-x_T',\dots,-x_T',-2x_T',-x_T',\dots,-x_T')'$.
Thus we can calculate $d-1$ differences $r_i$, assign $r_d=0$, and then apply the substitution function from proposition~\ref{prop:main} to get predictions. The resulting algorithm is Algorithm 1. We will further call it mAAR (multi-dimensional Aggregating Algorithm for Regression).

\begin{algorithm}[ht]
  \caption{mAAR for the Brier game}
  \label{alg:SAA}
  \begin{algorithmic}
    \STATE Fix $n$, $a>0$. $C=0, h=0$.
    \FOR{$t=1,2,\dots$}
      \STATE Read new $x_t \in \mathbf{X}$.
      \STATE $C = C + x_t x_t'$, $A = aI + \begin{pmatrix}
                                                2C & \cdots & C\\
                                                \vdots & \ddots & \vdots \\
                                                C  & \cdots & 2C\\
                                            \end{pmatrix}$
      \STATE Set $b_i = h + (x_t',\ldots,x_t',0,x_t',\ldots,x_t')'$, where $0$ is a zero-vector from $\mathbb{R}^n$ is placed at $i$-th position,\enspace $i=1,\ldots,d-1$.
      \STATE Set $z_i = (-x_t',\ldots,-x_t',-2x_t',-x_t',\ldots,-x_t')'$, where $-2x_t'$ is placed at $i$-th position,\enspace $i=1,\ldots,d-1$.
      \STATE Calculate $r_i := -b_i'A^{-1}z_i, r_d := 0,\enspace i=1,\ldots,{d-1}$.
      \STATE Solve $\sum_{i=1}^d(s-r_i)^+=2$ in $s\in\mathbb{R}$.
      \STATE Set $\gamma_t^i := (s-r_i)^+/2$, $\omega\in\Omega,\enspace i=1,\ldots,d$.
      \STATE Output prediction $\gamma_t\in\PPP(\Omega)$.
      \STATE Read observation $y_t$.
      \STATE $h_i = h_i - 2(y_t^i - y_t^d)x_t, h = (h_1',\ldots,h_{d-1}')'$.
    \ENDFOR
  \end{algorithmic}
\end{algorithm}

\subsection{Component-wise algorithm}\label{ssec:componentwisealg}
In this section we derive the component-wise algorithm. It gives predictions for each component of the outcome separately, and then combines them in a special way.

First we explain why we should not directly use the algorithm and the theoretical bound proposed in \cite{VovkCOS}. Vovk's experts do not allow us to take advantage of the fact that only one outcome is possible to happen at each moment. They are more suitable for the case when each input vector $x$ can belong to many classes simultaneously in case of classification. In other words, they are centered around the center $1/2$ of the prediction interval $[0,1]$: $\xi_i = 1/2 + \alpha_i x$. Assume that the number of outcomes is very large and the distribution on experts is normal $N(0,\sigma^2)$ with small $\sigma$. Then the average experts' prediction is $(1/2,\ldots,1/2,1-(d-1)/2))$, and the average loss of the experts on trials with the same outcome $y = y\{i\}$ (we can take $y = (1,0,\ldots,0)$) is $(d-1)/2^2 + (d-1)^2/2^2$. Components of experts \eqref{eq:experts1} concentrate around the point $1/d$, and so experts have the average loss $(d-1)/d^2 + (1-1/d)^2$. This loss is smaller than the loss of Vovk's experts for large values of $d$.

Our component-wise experts are expressed by
\begin{equation}\label{eq:experts2}
\xi_t^i = 1/d + \alpha_i' x_t, \quad i=1,\ldots,d.
\end{equation}
The derivation of the component-wise algorithm (further cAAR stands for component-wise Aggregating Algorithm Regression) is similar to the derivation of Algorithm~\ref{alg:SAA} for two outcomes. The initial distribution on each component of experts~\eqref{eq:experts2} is given by
\begin{equation*}
(a\tilde \eta/\pi)^{n/2}e^{-a \tilde \eta\|\alpha_i\|^2}d\alpha_i.
\end{equation*}
Note that the value for $\tilde \eta$ here will be different from $1$ since the loss function by each component is half of the Brier loss $\lambda(y,\gamma) = (y-\gamma)^2 + (1-y - (1-\gamma))^2$. We will further see that $\tilde \eta=2$.
The loss of expert $\xi(\alpha_i)$ over the first $T$ trials is
\begin{equation*}
\sum_{t=1}^T(y_t^i-1/d-\alpha'_ix_t)^2 = \alpha'_i\left(\sum_{t=1}^T x_tx_t'\right)\alpha_i - 2\alpha'_i\left(\sum_{t=1}^T (y_t^i-1/d)x_t\right) + \sum_{t=1}^T (y_t^i-1/d)^2.
\end{equation*}
Instead of the substitution function from Proposition~\ref{prop:main} we use the substitution function suggested in \citet{VovkCOS} for the one-dimensional game:
\begin{equation*}
\gamma_T^i = \frac{1}{2} + \frac{g_T(0)-g_T(1)}{2}
\end{equation*}
Therefore, the substitution function can be represented as
\begin{align}
\gamma_T^i &= \frac{1}{2} + \frac{1}{2}\log_{\tilde\beta} \frac{\tilde\beta^{g_T(0)}}{\tilde\beta^{g_T(1)}}\notag\\
&= \frac{1}{2} + \frac{1}{2}\log_{\tilde\beta} \frac{\int_{\mathbb{R}^n} e^{-\tilde \eta \alpha'_iB\alpha_i +
2\tilde\eta\alpha'_i\left(E  + (0-1/d)x_T \right)
-
\tilde\eta\left(W + 1/d^2\right)}
d\alpha_i}
{
\int_{\mathbb{R}^n} e^{-\tilde \eta \alpha'_iB\alpha_i +
2\tilde\eta\alpha'_i\left(E + (1-1/d)x_T\right)
-
\tilde\eta\left(W + (1-1/d)^2\right)}
d\alpha_i
} \notag\\
&= \frac{1}{d} + \frac{1}{2}F\left(B,-2E - \frac{d-2}{d}x_T,x_T\right) \notag\\
&= \frac{1}{d} + \left(\sum_{t=1}^{T-1} (y_t^i-1/d)x_t' + \frac{d-2}{2d}x_T'\right)\left(aI + \sum_{t=1}^T x_tx_t'\right)^{-1}x_T \label{eq:cAAR}
\end{align}
for $i=1,\ldots,d$.
Here $B = aI + \sum_{t=1}^T x_tx_t'$, $E = \sum_{t=1}^{T-1} (y_t^i-1/d)x_t$, $W=\sum_{t=1}^{T-1} (y_t^i-1/d)^2$, $\tilde \beta = e^{-\tilde \eta}$. The transitions are justified using Lemma~\ref{lem1} and Lemma~\ref{lem2}.

Then this method projects its prediction onto the prediction simplex such that the loss does not increase. We use the projection algorithm suggested in~\citet{Michelot1986}.

\begin{algorithm}[ht]
  \caption{Projection of a point from $\R^n$ onto probability simplex.}
  \label{alg:proj}
  \begin{algorithmic}
    \STATE Initialize $I = \emptyset$, $x =\textbf{1} \in \R^d$.
    \STATE Let $\gamma_T$ be the prediction vector and $|I|$ is the dimension of the set $I$.
    \WHILE{$1$}
      \STATE $\gamma_T = \gamma_T - \frac{\sum_{i=1}^d \gamma_T^i-1}{d-|I|}$;
      \STATE $\gamma_T^i = 0, \forall i \in I$;
      \STATE If $\gamma_T^i \ge 0$ for all $i = 1,\ldots,d$ then break;
      \STATE $I = I \bigcup \{i: \gamma_T^i < 0\}$;
      \STATE If $\gamma_T^i < 0$ for some $i$ then $\gamma_T^i = 0$;
    \ENDWHILE
  \end{algorithmic}
\end{algorithm}

\section{Theoretical bound}\label{sec:linearbound}
We derive the theoretical bounds for the losses of Algorithm~\ref{alg:SAA} and of a naive component-wise algorithm predicting in the same framework.
\subsection{Component-wise algorithm}\label{ssec:componentwise}
We prove here the theoretical bound for the loss of cAAR. The following lemma is the main tool helping us to prove our theorems. It is easy to prove the following statement (Lemma 1 from \cite{VovkCOS}):
\begin{lemma}\label{lemV}
  If the learner follows the Aggregating Algorithm in a perfectly mixable game, then for every positive integer $T$, every sequence of outcomes of the length $T$, and any initial weights distribution on experts $P_0(d\alpha)$ it suffers loss satisfying for any $\alpha \in \Theta$
  \begin{equation}\label{eq:AAloss}
    L_T(\mathrm{AA}(\eta,P_0)) \le \log_\beta \int_{\Theta} \beta^{L_T(\alpha)} P_0(d\alpha).
  \end{equation}
\end{lemma}
\BP
We proceed by induction in $T$: for $T=0$ the inequality is obvious, and for $T>0$ we have:
\begin{multline*}
    L_T(\mathrm{AA}(\eta,P_0))
    \le
    L_{T-1}(\mathrm{AA}(\eta,P_0))
    +
    g_T(\omega_T) \\
    =
    \log_\beta \int_{\Theta} \beta^{L_{T-1}^\theta} P_0(d\theta)
    +
    \log_\beta \int_{\Theta} \beta^{\lambda(\omega_T,\xi_t^\theta)}
    \frac{\beta^{L_{T-1}^\theta}}{\int_{\Theta} \beta^{L_{T-1}^\theta} P_0(d\theta)}P_0(d\theta)\\
    =
    \log_\beta \int_{\Theta} \beta^{L_T^\theta} P_0(d\theta) \enspace.
  \end{multline*}
  Here the second equality follows from the inductive assumption, the definition~\eqref{eq:genpred} of $g_T$, and~\eqref{eq:wupdate}.
\EP
The loss of the component-wise algorithm by one component is bounded as in the following theorem.

\begin{theorem}\label{thm:losscompon}
Let the outcome space in the prediction game be $[A,B], A,B \in \mathbb{R}$. Assume experts' predictions at each step are $\xi_t = C + \alpha' x_t$, where $\alpha \in \mathbb{R}^n$, $C \in \mathbb{R}$ is the same for all the experts $\alpha$, and $\|x_t\|_\infty \le X,\forall t$. There exists a prediction algorithm producing $\gamma_i \in \R, i=1,\ldots,d$ such that for any $a > 0$, every positive integer $T$, every sequence of input vectors and outcomes of the length $T$ and any $\alpha \in \R^n$ we have
\begin{equation}\label{eq:losscompon}
\sum_{t=1}^T (\gamma_t-y_t)^2 \le \sum_{t=1}^T (\xi_t-y_t)^2 + a\|\alpha\|^2_2
 + \frac{n(B-A)^2}{4} \ln\left(\frac{TX^2}{a}+1\right).
\end{equation}
\end{theorem}
\BP
We need to prove that the game is perfectly mixable (see \eqref{eq:lAA}) and find the optimal parameter $\eta$ for the algorithm. Implications similar to the ones in the proof of Lemma 2 from \cite{VovkCOS} lead to the inequality $\eta \le \frac{2}{(B-A)^2}$. Clearly, Lemma \ref{lemV} holds for our case, so we need only to calculate the difference between the right-hand side of \eqref{eq:AAloss}
\begin{multline*}
\log_\beta \int_{\mathbb{R}^n} d\alpha(a\eta/\pi)^{n/2}\exp\left[-\eta\alpha'\left(aI + \sum_{t=1}^T x_tx_t'\right)\alpha\right.\\
+ \eta\left.2\alpha'\left(\sum_{t=1}^T (y_t-C)x_t\right) -\eta\sum_{t=1}^T (y_t-C)^2\right].
\end{multline*}
and the loss of the best expert $\alpha_0'\left(aI + \sum_{t=1}^T x_tx_t'\right)\alpha_0
- 2\alpha_0'\left(\sum_{t=1}^T (y_t-C)x_t\right) + \sum_{t=1}^T (y_t-C)^2.$
Here $\alpha_0$ is the point where the minimum of the quadratic form is attained. Then due to Lemma \ref{lem1} this difference will be equal to
\begin{equation*}
  \frac{1}{2\eta}\ln\det \left( I+\frac{1}{a}\sum_{t=1}^T x_tx_t'\right)
 \le \frac{n(B-A)^2}{4}\ln\left(\frac{TX^2}{a}+1\right).
\end{equation*}
We bound the determinant of a symmetric positive definite matrix by the product of its diagonal elements (see \cite{Beckenbach1961}, Chapter 2, Theorem 7) and use $\eta = \frac{2}{(B-A)^2}$.
\EP

Interestingly, the theoretical bound for the regression algorithm depends only on the size of the prediction interval but not on the location of it. It also does not depend on the concentration point of experts. We use the component-wise algorithm to predict each component separately.
\begin{theorem}\label{thm:componwise}
  If $\|x_t\|_\infty \le X, \forall t,$ then for any $a > 0$, every positive integer $T$, every sequence of outcomes of the length $T$, and any $\alpha \in \R^{n(d-1)}$ the loss $L_T$ of the component-wise algorithm satisfies
  \begin{equation}\label{eq:regretstraight}
    L_T \le L_T(\alpha) + da\|\alpha\|^2_2 + \frac{nd}{4}\ln\left(\frac{TX^2}{a}+1\right).
  \end{equation}
\end{theorem}
\BP
We extend the class of experts in \eqref{eq:experts1} in~\eqref{eq:experts2}. The algorithm predicts each component of the outcome separately. Summing theoretical bounds \eqref{eq:losscompon} for $d$ components of the outcome, taking $\alpha_d = - \sum_{i=1}^{d-1}\alpha_i'$, and using the Cauchy inequality $\|\sum_{i=1}^{d-1}\alpha_i\|^2_2 \le (d-1)\sum_{i=1}^{d-1}\|\alpha_i\|^2_2$ we get the bound. To give probability forecasts we can project prediction points on the prediction simplex using Algorithm~\ref{alg:proj}. The bound will then hold by Lemma~\ref{lem:projection}.
\EP

\subsection{Linear forecasting}
The theoretical bound for the loss of the Algorithm~\ref{alg:SAA} is
\begin{theorem}\label{thm:main}
  If $\|x_t\|_\infty \le X, \forall t,$ then for any $a > 0$, every positive integer $T$, every sequence of outcomes of the length $T$, and any $\alpha \in \R^{n(d-1)}$ mAAR$(2a)$ satisfies
  \begin{equation}\label{eq:main}
    L_T(\mathrm{mAAR(}2a\mathrm{)}) \le L_T(\alpha) + 2a\|\alpha\|^2_2
    + \frac{n(d-1)}{2}\ln\left(\frac{TX^2}{a}+1\right).
  \end{equation}
\end{theorem}
\begin{proof}
We apply mAAR with the parameter $b=2a$. Recall that $C = \sum_{t=1}^T x_t x_t'$. Following the line of the proof of Theorem~\ref{thm:losscompon} with $\eta=1$ we get the theoretical bound.
\EP
We can derive a slightly better theoretical bound: in the determinant of $A$ one should subtract the second block raw from the first one and then add the first block column to the second one, then repeat this $d-2$ times.
\begin{proposition}\label{prop:betterbound}
In the conditions of Theorem~\ref{thm:main} mAAR($a$) satisfies
  \begin{multline}\label{eq:betterbound}
    L_T(\mathrm{mAAR(}a\mathrm{)}) \le L_T(\alpha) + a\|\alpha\|^2_2\\
    + \frac{n(d-2)}{2}\ln\left(\frac{TX^2}{a}+1\right) + \frac{n}{2}\ln\left(\frac{TX^2d}{a}+1\right).
  \end{multline}
\end{proposition}
The theoretical bound~\eqref{eq:main} is worse asymptotically by $d$ than the bound~\eqref{eq:regretstraight} of the component-wise algorithm, but it is better in the beginning, especially when the norm of the best expert $\|\alpha\|$ is large. This can happen in the important case when the dimension of the input vector is larger than the size of the prediction set: $n >> T$.
\section{Kernelization}\label{sec:kernel}
In some cases the linear model can be considered not rich enough to describe data well, and a more complicated model is needed. We use a popular in computer learning kernel trick, firstly applied to the AAR in \citet{Gammerman2004}. We derive an algorithm competing with all sets of functions from an RKHS with $d-1$ elements.

\subsection{Derivation of the algorithm}
\begin{definition}{\rm
Let us take $x_1,\ldots,x_n \in \mathbf{X}$. A \emph{kernel function} is a nonnegative function $K: \R^n \times \R^n \to \R$ satisfying $\sum_{i,j=1}^n K(x_i,x_j)\xi_i\xi_j \ge 0$ for all positive integers $n$, all $x_1,\ldots,x_n \in \mathbf{X}$, and $\xi_1,\ldots,\xi_n \in \mathbb{R}$.
}\end{definition}
An RKHS contains all linear regressors $\langle\Phi(\cdot),h\rangle_H$ defined by means of a feature map (for all the definitions see \citealp{Scholkopf2002}). It can also be defined in a different equivalent way as a functional Hilbert space with continuous evaluation functional $\varphi: f \in \mathcal{F} \mapsto f(x)$ for each $x \in \mathbf{X}$. We will use the notation $c_\mathcal{F}(x)$ for the norm of this functional: $c_{\mathcal{F}}(x) := \sup_{f:\|f\|_{\mathcal{F}} \le 1} |f(x)|$ and for the embedding constant $c_{\mathcal{F}} := \sup_{x \in \mathbf{X}} c_{\mathcal{F}}(x)$ and assume $c_{\mathcal{F}} < \infty$.

Our algorithm competes with the following experts:
\begin{align}\label{eq:expertsK}
\xi^1_t &= 1/d + f_1(x_t)\notag\\
&\ldots \notag\\
\xi^{d-1}_t &= 1/d + f_{d-1}(x_t)\\
\xi^d_t &= 1 - \xi^1 - \dots - \xi^{d-1}. \notag
\end{align}
Here $f_1,\ldots,f_{d-1} \in \mathcal{F}$ are any functions from some RKHS $\mathcal{F}$. We start by rewriting mAAR in the dual form. Denote
\begin{align*}
    \widetilde{Y}_i &=-2(y_1^i-y_1^d,\ldots,y_{T-1}^i-y_{T-1}^d,-1/2),\\
    \overline{Y}_i &=-2(y_1^i-y_1^d,\ldots,y_{T-1}^i-y_{T-1}^d,0)\\
    \widetilde{k}(x_T) & =(x_1'x_T,\ldots,x_T'x_T)',\\
    \widetilde{K} &=(x_s',x_t)_{s,t} \mbox{ is the matrix of scalar products}
\end{align*}
for $i=1,\ldots,d-1$, $s,t=1,\ldots,T$.
We show that the predictions of mAAR can be represented in terms of variables defined above. We will need the following matrix property.
\begin{proposition}\label{prop:matrixequal}
Let $B,C$ be matrices such that the number of rows in $B$ equals to the number of columns in $C$, and identity matrices $I$. If $aI+CB$ and $aI+BC$ are nonsingular then
\begin{equation}\label{eq:matrixequal}
B(aI+CB)^{-1}=(aI+BC)^{-1}B.
\end{equation}
\end{proposition}
\BP
This is equivalent to $(aI+BC)B=B(aI+CB)$. That is true because of distributivity of matrix multiplication.
\EP
Let us set $A = \left(aI + \begin{pmatrix}
                        2\widetilde{K} & \cdots & \widetilde{K}\\
                        \vdots         & \ddots & \vdots \\
                        \widetilde{K}  & \cdots & 2\widetilde{K}\\
              \end{pmatrix}\right)$.
\begin{lemma}\label{lemK}
  On trial $T$ values $r_i$ for $i=1,\ldots,d-1$ in mAAR can be represented as
  \begin{multline}\label{eq:ker}
   r_i= \begin{pmatrix}
            \widetilde{Y}_1 & \cdots & \overline{Y}_i & \cdots & \widetilde{Y}_{d-1}
        \end{pmatrix}\\
   \cdot A^{-1}
         \begin{pmatrix}
                        \widetilde{k}(x_T)' &
                        \cdots &
                        2\widetilde{k}(x_T)' &
                        \cdots &
                        \widetilde{k}(x_T)'
          \end{pmatrix}'.
  \end{multline}
\end{lemma}
\BP
By $M=(x_1,\ldots,x_T)$ denote a matrix $n \times T$ of column input vectors. Let us set
\begin{equation*}
B = \begin{pmatrix}
    2M & \cdots & M\\
    \vdots  & \ddots & \vdots \\
    M  & \cdots & 2M\\
    \end{pmatrix},
C = \begin{pmatrix}
    M' & \cdots  & 0\\
    \vdots  & \ddots & \vdots \\
    0  & \cdots & M'\\
    \end{pmatrix}.
\end{equation*}
Then $h_i$ from the algorithm mAAR equals $h_i=M\overline{Y}_i \in \mathbb{R}^{n}$.
Decompose
$
b_i'= \begin{pmatrix}
    \widetilde{Y}_1 & \cdots & \overline{Y}_i & \cdots & \widetilde{Y}_{d-1}
    \end{pmatrix}
    C,
$
where only the $i$-th block uses $\overline{Y}_i$.
The matrix $A$ is equal
$
A = aI + BC.
$
Using proposition \ref{prop:matrixequal}
\begin{multline*}
r_i = -b_i'A^{-1}z_i =-
                 \begin{pmatrix}
                    \widetilde{Y}_1 & \cdots & \overline{Y}_i & \cdots & \widetilde{Y}_{d-1}
                 \end{pmatrix}\\
\cdot(aI + CB)^{-1} C
                \begin{pmatrix}
                        -x_T' & \cdots & -2x_T' & \cdots -x_T'
                \end{pmatrix}'.
\end{multline*}
Note that $\widetilde{K}=M'M$ and $\widetilde{k}(x_T)=M'x_T$, thus \eqref{eq:ker} holds.
\EP
If instead of dot product in $\widetilde{K},\widetilde{k}(x_T)$ we can choose a different kernel (classical examples of kernels are Gaussian (RBF): $K(x_i,x_j)=e^{-\frac{\|x_i-x_j\|^2}{2\sigma^2}}$, Vapnik's polynomial $K(x_i,x_j)=(x_i \cdot x_j+1)^d$, etc.).
To get predictions one can use the same substitution function from Proposition~\ref{prop:main}. We call this algorithm mKAAR (K for Kernelized).

\subsection{Theoretical bound for the kernelized algorithm}
To derive a theoretical bound for the loss of mKAAR we will use the following matrix determinant identity lemma.
\begin{lemma}[Matrix determinant identity]\label{lemIden}
Let $B,C$ are as in Proposition~\ref{prop:matrixequal}, and $a$ is a real number. Then
$
\det (aI + BC)=\det (aI + CB).
$
\end{lemma}
\BP
The proof is by considering a block matrix identity.
\EP
The main theorem follows from the property of RKHS called Representer theorem \citep[see][Theorem 4.2]{Scholkopf2002}.
\begin{theorem}[Representer theorem]\label{thm:repres}
Denote by $g:[0,\infty) \to \mathbb{R}$ a strictly monotonic increasing function. Assume $\mathbf{X}$ is an arbitrary set, and $\mathcal{F}$ is a Reproducing Kernel Hilbert Space of functions on $\mathbf{X}$ with the given kernel $K:\mathbf{X}^2 \to \mathbb{R}$. Assume we also have a positive integer $T$ and an arbitrary loss function $c: (\mathbf{X} \times \mathbb{R}^2)^T \to \mathbb{R} \bigcup \{\infty\}$. Then each minimizer $f \in \mathcal{F}$ of
\begin{equation*}
c\left( (x_1,y_1,f(x_1)),\ldots,(x_T,y_T,f(x_T))\right) + g(\|f\|_{\mathcal{F}})
\end{equation*}
admits a representation of the form $f(x) = \sum_{i=1}^T \alpha_i K(x_i,x)$ for any $x \in \mathbf{X}$ and reals $\alpha_i, i=1,\ldots,T$.
\end{theorem}
The theoretical bound for the loss of mKAAR is proven in the following theorem.
\begin{theorem}\label{thm:mainK}
Assume $\mathbf{X}$ is an arbitrary set of inputs and $\mathcal{F}$ is a Reproducing Kernel Hilbert Space of functions on $\mathbf{X}$ with the given kernel $K:\mathbf{X}^2 \to \mathbb{R}$. Then for any $a > 0$, any $f_1,\ldots,f_{d-1} \in \mathcal{F}$, any positive integer $T$, and any sequence of inputs and outputs $(x_1,y_1),\ldots,(x_T,y_T)$
\begin{equation}\label{eq:BoundmKAAR2}
L_T(\mathrm{mKAAR}) \le L_T(f) + a\sum_{i=1}^{d-1}\|f_i\|^2_{\mathcal{F}}
                    + \frac{1}{2}\ln\det A
\end{equation}
Here the matrix $\widetilde{K}$ is a matrix of kernel values $K(x_i,x_j)$, $i,j=1,\ldots,T$.
\end{theorem}
\BP
The bound follows from Theorem \ref{thm:main} for mAAR and the Representer theorem. Let us first consider the case with scalar product kernel. Denote $C = \sum_{t=1}^T x_t x_t '$. By Lemma~\ref{lemIden} and calculations similar to ones in the proof of Lemma~\ref{lemK} we have the equality of determinants.
So we can use any other kernel instead of scalar product to get the term with the determinant.
The Representer theorem assures that the minimum of the expression $L_T(f) + a\sum_{i=1}^{d-1}\|f_i\|^2_{\mathcal{F}}$ by $f$-s is reached on a linear regressor.
\EP
We can represent the bound (\ref{eq:BoundmKAAR2}) in another form which is more familiar from the on-line prediction literature:
\begin{corollary}\label{cor:Bound}
Under assumptions of Theorem~\ref{thm:mainK} and if we know the number of steps $T$ in advance and are given $F > 0$, the mKAAR reaches the performance
\begin{equation}\label{eq:BoundmKAAR3}
L_T(\mathrm{mKAAR}) \le L_T(f) + 2c_{\mathcal{F}}F\sqrt{(d-1)T},
\end{equation}
for any $f_1,\ldots,f_{d-1} \in \mathcal{F}: \sum_{i=1}^{d-1}\|f_i\|^2_{\mathcal{F}} \le F$.
\end{corollary}
\BP
Bounding the logarithm of the determinant we have
$
\ln\det A \le (d-1)T \ln \left(1+\frac{2c^2_{\mathcal{F}}}{a}\right).
$
We can choose the value for $a$ where the minimum is achieved: $a=\frac{c_{\mathcal{F}}\sqrt{(d-1)T}}{F}$.
\EP

\section{Experiments}\label{sec:experiments}
We run our algorithms on six real world time-series data sets. In the time series we consider there are no signals attached to the outcomes. However we can take vectors consisting of previous observations (we shall take ten of those) and use them as signals. Data set DEC-PKT\footnote{Data sets can be found http://ita.ee.lbl.gov/html/traces.html.} contains an hour's worth of all wide-area traffic between Digital Equipment Corporation and the rest of the world. Data set LBL-PKT-4\footnotemark[\value{footnote}] consists of observations of another hour of traffic between the Lawrence Berkeley Laboratory and the rest of the world. We transformed both the data sets in such a way that each observation is the number of packets in the corresponding network during a fixed time interval of one second. The other four datasets\footnote{Data sets can be found http://www.neural-forecasting-competition.com/index.htm.} (C4,C9,E5,E8) relate to transportation data. Two of them (C9,C11) contain low-frequency monthly traffic measures. Two of them (E5,E8) contain high-frequency day traffic measures. On each of these data sets the following operations were performed: subtraction of the mean value and division by the maximum absolute value. The resulting time series are shown in Figure~\ref{fig:series}.

\begin{figure}
\center
\subfigure[DEC-PKT series]{\includegraphics[width=0.22\textwidth]{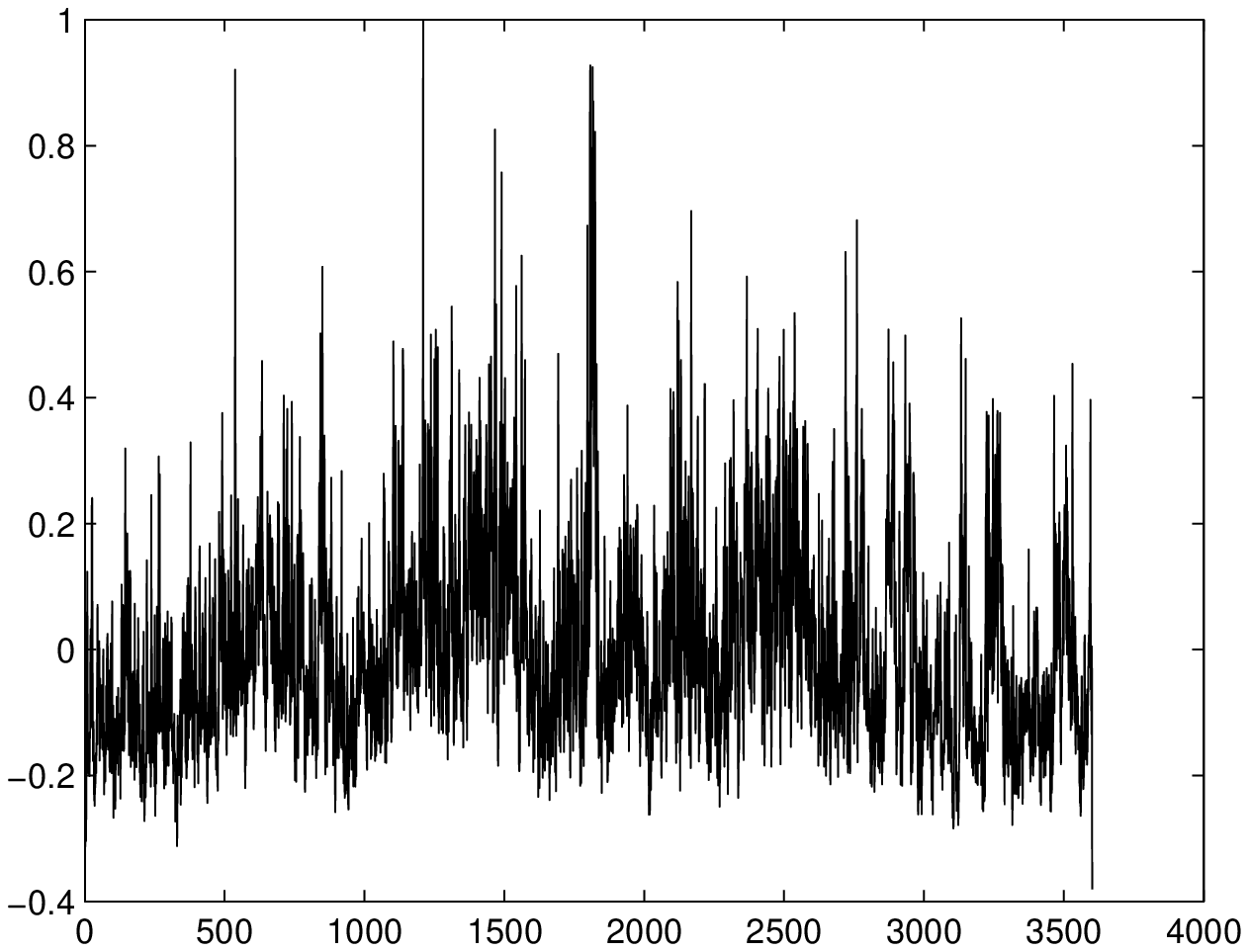}}
\subfigure[LBL-PKT series]{\includegraphics[width=0.22\textwidth]{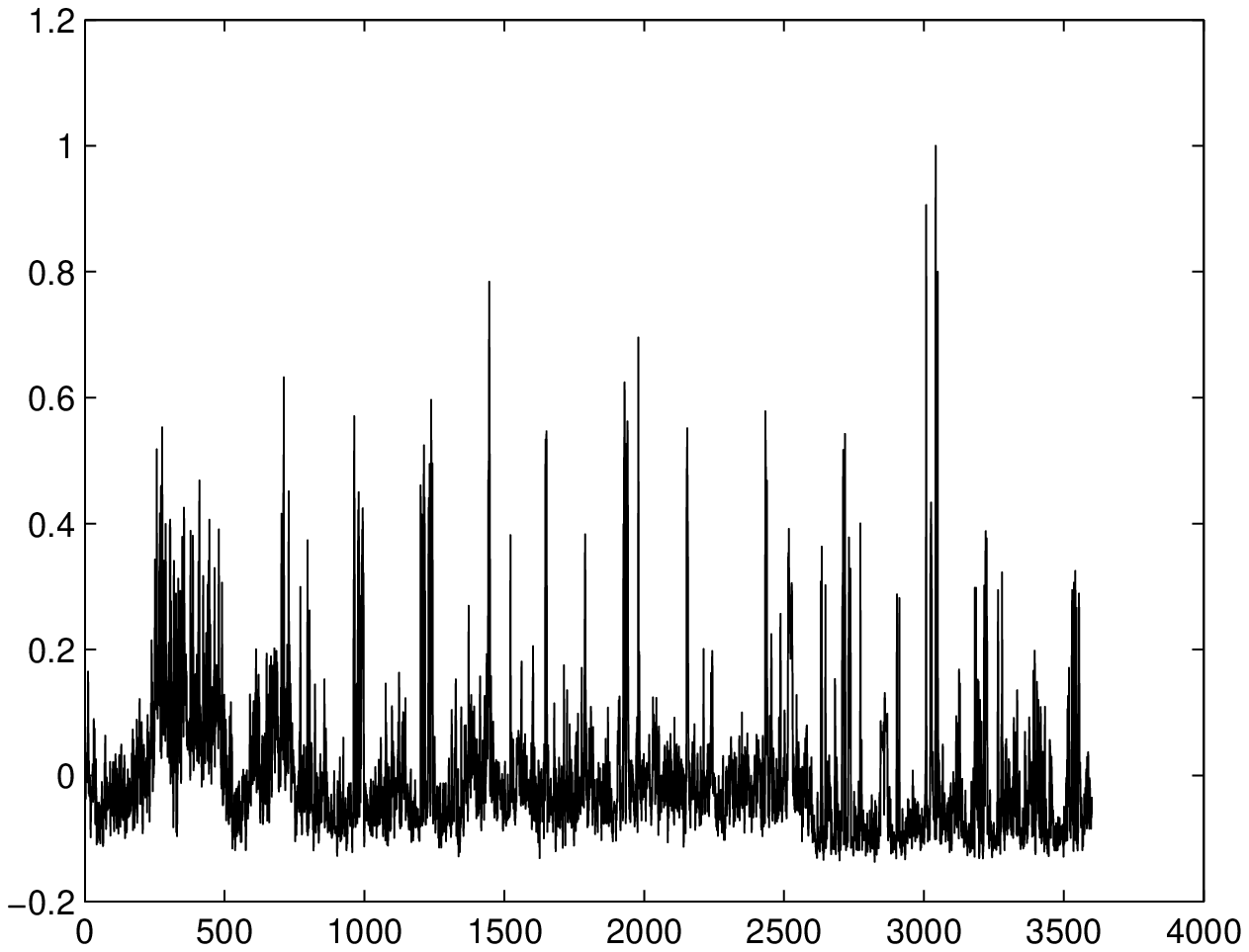}}
\subfigure[C4 series]{\includegraphics[width=0.22\textwidth]{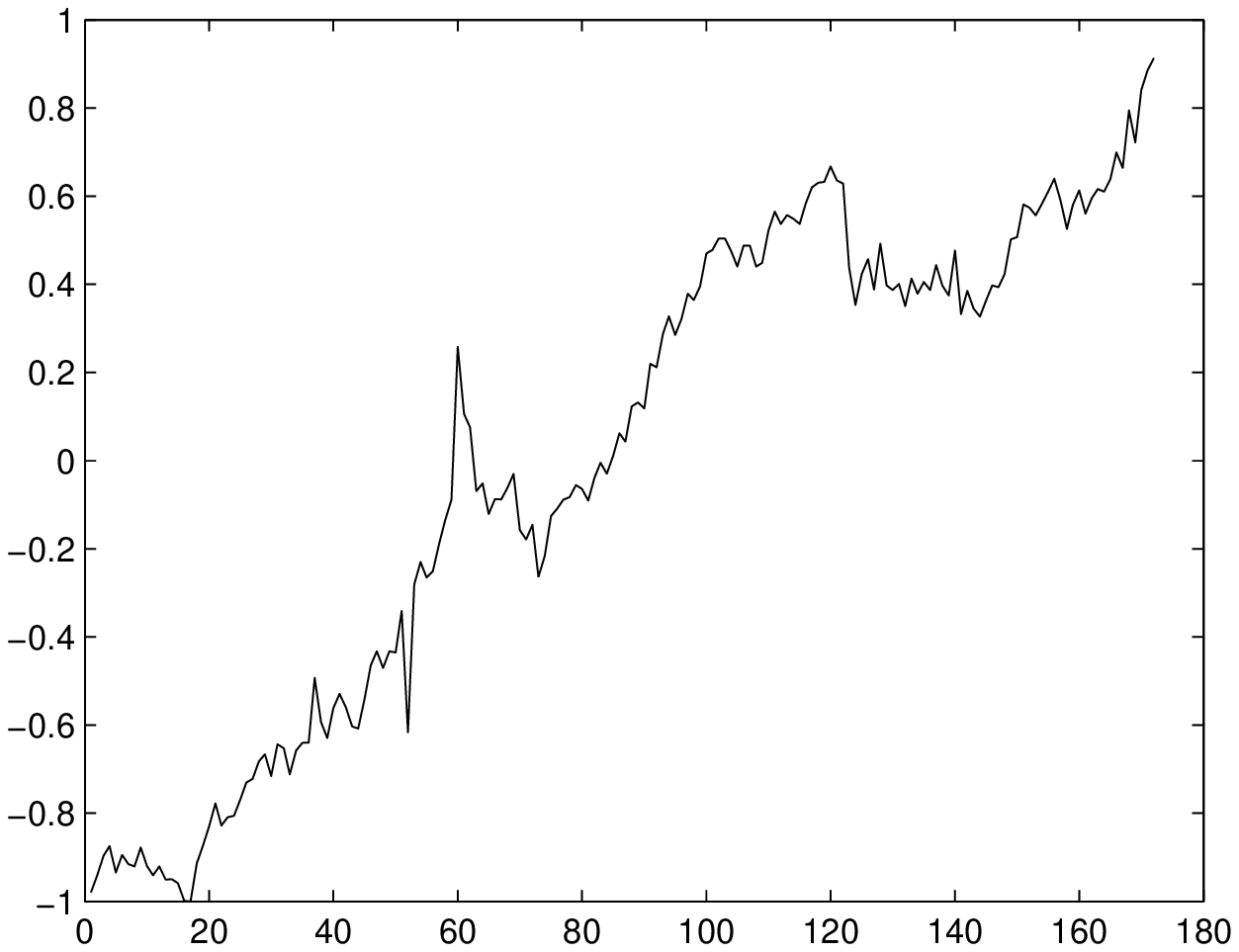}}
\subfigure[C9 series]{\includegraphics[width=0.22\textwidth]{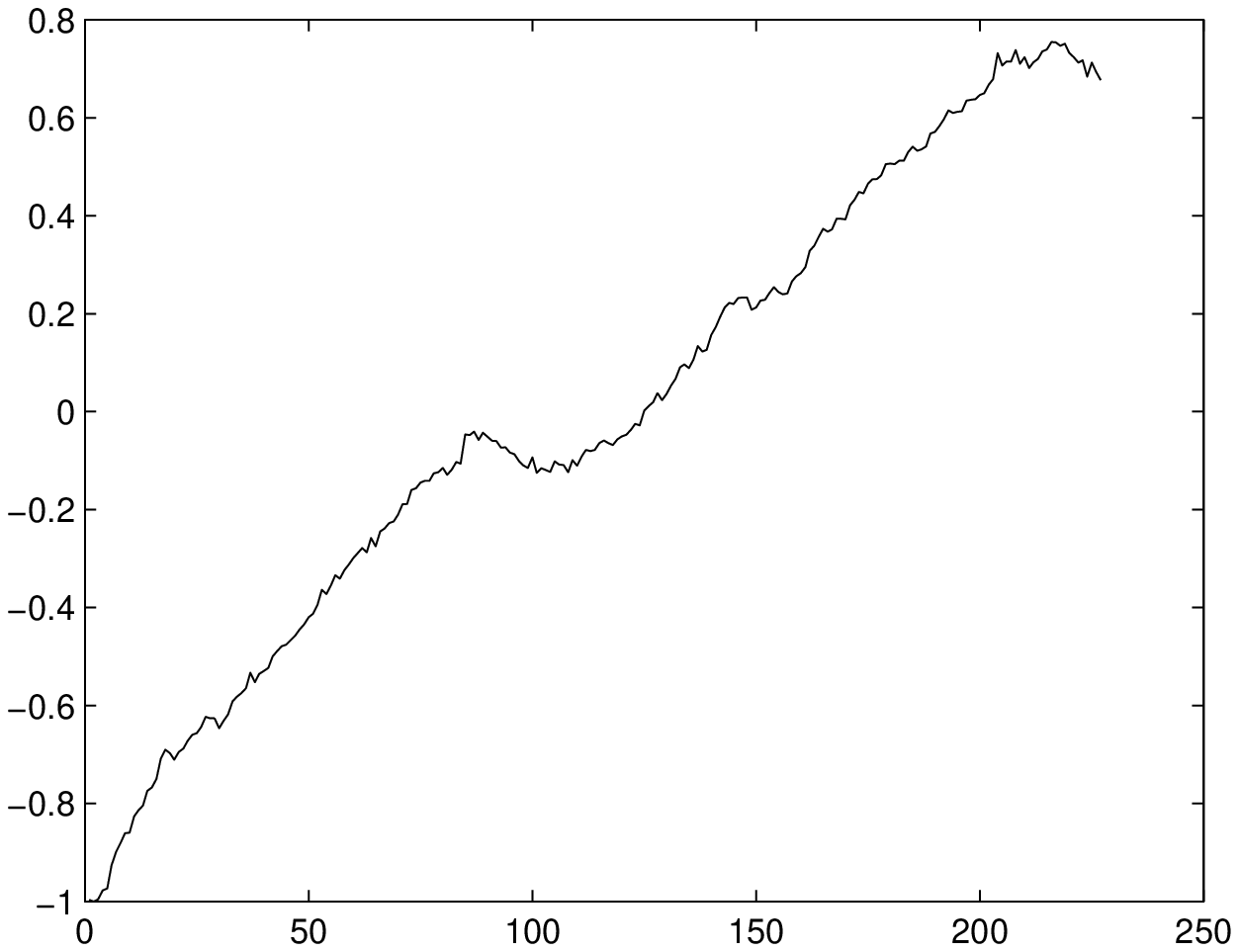}}
\subfigure[E5 series]{\includegraphics[width=0.22\textwidth]{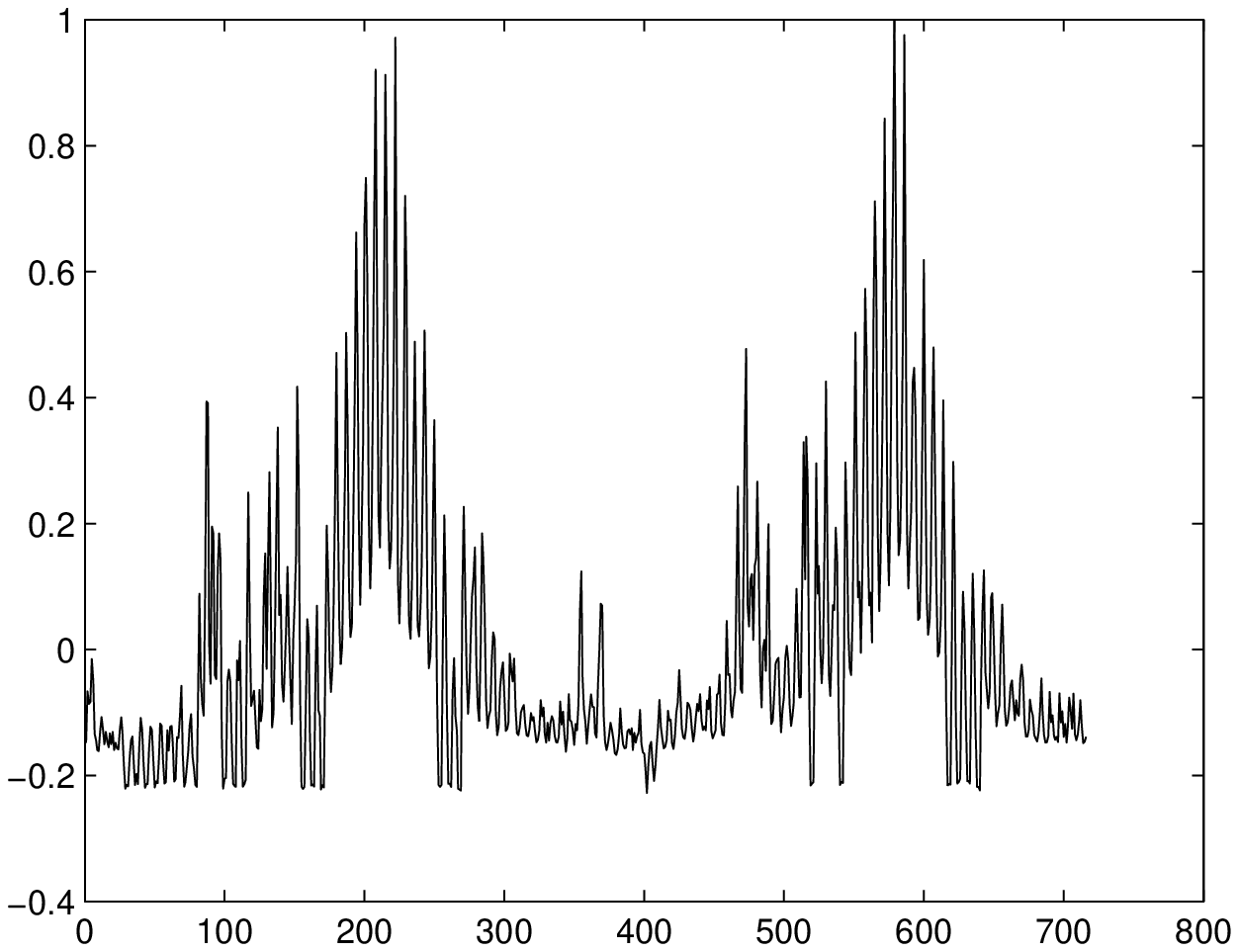}}
\subfigure[E8 series]{\includegraphics[width=0.22\textwidth]{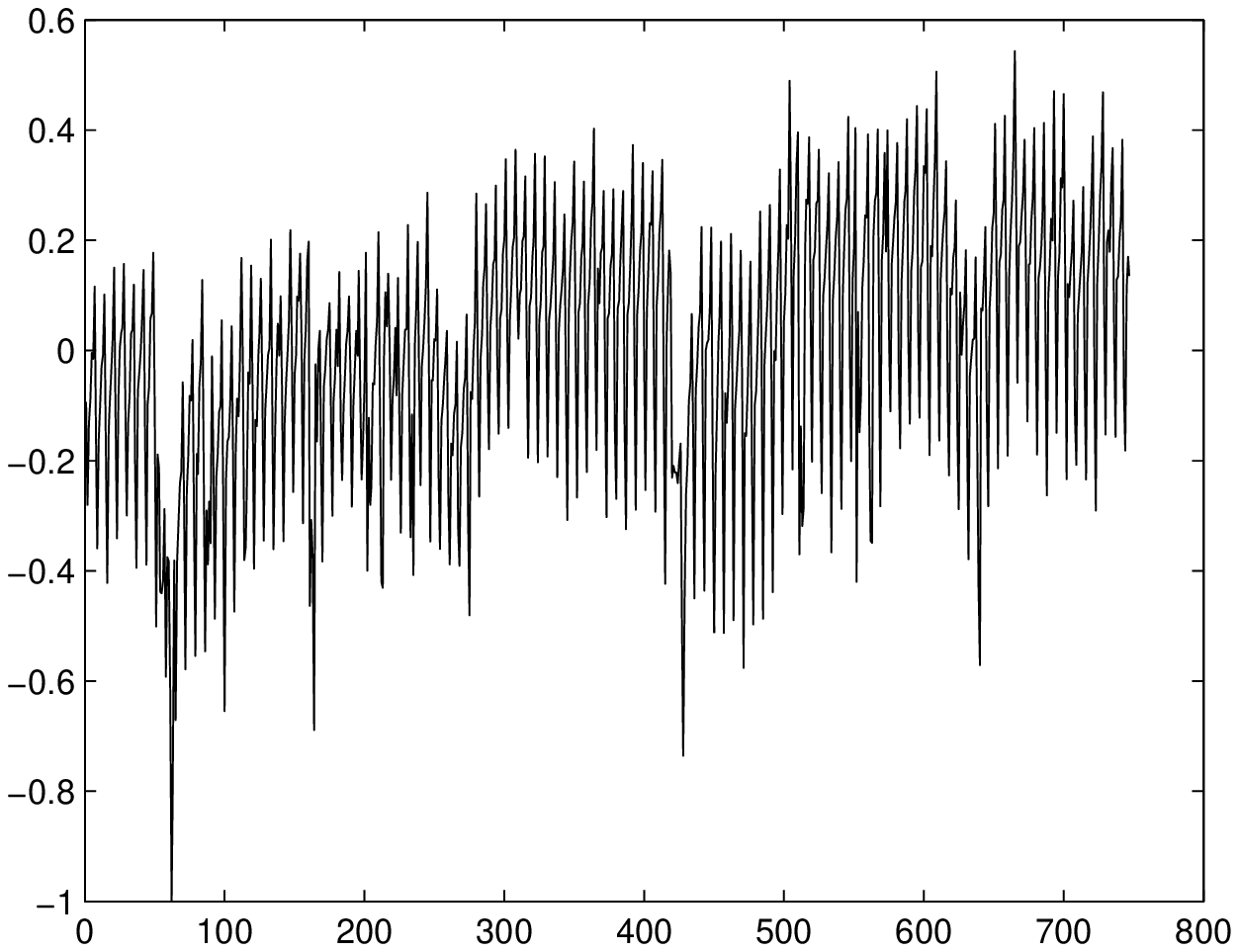}}
\caption{Time series from 6 data sets.}\label{fig:series}
\end{figure}

We used ten previous observations as an input vector for tested algorithms at each prediction step. We are solving the 3-class classification problem: we predict whether the next value in a time series will be more than the previous value plus a precision parameter $\epsilon$, less than that value, or lies in the $2\epsilon$ tube around the previous value. The precision $\epsilon$ is chosen to be the median of all the changes in a data set. In order to assess the quality of predictions, we calculate the cumulative square loss at the last two thirds of each time series (test set) and divide it by the number of examples (MSE). Since we are considering the online setting, we could calculate the cumulative loss from the beginning of each time series. However our approach is not sensitive to starting effects, it allows us to choose the ridge parameter $a$ fairly on the training set, and it allows us to compare the performance of our algorithms with batch algorithms, which would be normally used to solve this problem.

The square loss on the test set takes into account the quality of an algorithm only at the very end of the prediction process, and does not consider the quality during the process. We introduce another quality measure: at each step in the test set we calculate MSE of an algorithm until this step. After all the steps we average these MSEs (AMSE). Clearly, if one algorithm is better than another on the whole test set (its total MSE is smaller) but was often worse on many parts of the test set (total MSEs of many parts of the set is larger), this measure takes it into account.

We compare the performance of our algorithms with
the multinomial logistic regression (mLog),
because it is a standard classification algorithm
which gives probability predictions:
\begin{equation*}
\gamma_{\mathrm{mLog}}^i = \frac{e^{\theta^i x}}{\sum_{i=1}^d e^{\theta^i x}}
\end{equation*}
for all the components of the outcome $i=1,\ldots,d$. In our case $d=3$. Here parameters $\theta_1,\ldots,\theta_d$ are estimated from the training set. We apply this algorithm in two regimes: batch regime, where the algorithm learns only on the training set and is tested on the test set (and thus $\theta$ is not updated on the test set); and in the online regime, where at each step new parameters $\theta$ are found, and only one next outcome is predicted. The second regime is more fair to compare with online algorithms, but the first one is standard and faster. In both regimes logistic regression does not have theoretical guarantees on the square loss.

We also compare our algorithms
with the simple predictor
predicting the average of the ten previous outcomes
(and thus it always gives probability predictions).

We are not aware of other efficient algorithms for online probability prediction,
and thus logistic regression and simple predictor as the only baselines. Component-wise algorithms which could be used for online prediction (e.g., Gradient Descent, \citealt{Kivinen1997}, Ridge Regression, \citealt{Hoerl2000}), have to use normalization by Algorithm~\ref{alg:proj}. Thus they have to be applied in a different way than they are described in the corresponding papers, and can not be fairly compared with our algorithms.

The ridge for our algorithms is chosen to achieve the best MSE on the training set: the first third of each series. The results are shown in Table~\ref{tab:results}. We highlight the most precise algorithms for different data sets. We also show time needed to make predictions on the whole data set. The algorithms were implemented in Matlab R2007b and run on the laptop with 2Gb RAM and processor Intel Core 2, T7200, 2.00GHz.

As we can see from the table, online methods perform better than the batch method. Online logistic regression performs well, but is very slow. Our algorithms perform similar to each other and comparable to the online logistic regression, but are much faster.

\begin{table}[ht]
\begin{center}
\begin{tabular}{|c|c|c|c|}
  \hline
  \textbf{Set/Algorithm} & \textbf{MSE} & \textbf{AMSE} & \textbf{Time}\\
  \hline
  \multicolumn{4}{|l|}{DEC-PKT} \\
  \hline
  cAAR&	0.45906&	0.45822&	0.578	\\
  mAAR&	0.45906&	0.45822&	1.25	\\
  mLog&	0.46107&	0.46265&	0.375	\\
  mLog Online&	\textbf{0.45751}&	0.45762&	2040.141	\\
  Simple&	0.58089&	0.57883&	0	\\
  \hline
  \multicolumn{4}{|l|}{LBL-PKT} \\
  \hline
  cAAR&	0.48147&	0.479&	0.579	\\
  mAAR&	0.48147&	0.479&	1.266	\\
  mLog&	0.47749&	0.47482&	0.391	\\
  mLog Online&	\textbf{0.47598}&	0.47398&	2403.562	\\
  Simple&	0.57087&	0.5657&	0.016	\\
  \hline
  \multicolumn{4}{|l|}{C4} \\
  \hline
  cAAR&	0.64834&	0.65447&	0.015	\\
  mAAR&	\textbf{0.64538}&	0.65312&	0.062	\\
  mLog&	0.76849&	0.77797&	0.016	\\
  mLog Online&	0.68164&	0.7351&	4.328	\\
  Simple&	0.69037&	0.69813&	0.016	\\
  \hline
  \multicolumn{4}{|l|}{C9} \\
  \hline
  cAAR&	\textbf{0.63238}&	0.64082&	0.015	\\
  mAAR&	0.63338&	0.64055&	0.063	\\
  mLog&	0.97718&	0.91654&	0.031	\\
  mLog Online&	0.71178&	0.75558&	10.625	\\
  Simple&	0.6509&	0.65348&	0	\\
  \hline
  \multicolumn{3}{|l|}{E5} \\
  \hline
  cAAR&	0.34452&	0.34252&	0.078	\\
  mAAR&	0.34453&	0.34252&	0.219	\\
  mLog&	0.31038&	0.30737&	1.109	\\
  mLog Online&	\textbf{0.30646}&	0.30575&	446.578	\\
  Simple&	0.58212&	0.58225&	0	\\
  \hline
  \multicolumn{3}{|l|}{E8} \\
  \hline
  cAAR&	0.29395&	0.29276&	0.078	\\
  mAAR&	0.29374&	0.29223&	0.25	\\
  mLog&	0.31316&	0.30382&	0.109	\\
  mLog Online&	\textbf{0.27982}&	0.27068&	83.125	\\
  Simple&	0.69691&	0.70527&	0.016	\\
  \hline
\end{tabular}
\end{center}
\caption{The square losses and prediction time (sec) of different algorithms applied for time series prediction. cAAR and mAAR state for the derived algorithms, mLog states for the logistic regression, mLogOnline states for online logistic regression, and Simple stands for the simple average predictor.}
\label{tab:results}
\end{table}

\section{Discussion}\label{sec:conclusion}
We consider an important generalization of the online classification problem. We presented new algorithms which give probability predictions in the Brier game. Both algorithms do not involve any numerical integration, and can be easily computed. Both algorithms have theoretical guarantees on their cumulative losses. One of the algorithms is kernelized and a theoretical bound is proven for the kernelized algorithm. We performed experiments with linear algorithms and showed that they perform relatively well. We compared them with the logistic regression: the benchmark algorithm giving probability predictions.

Competing with linear experts in the case where possible outcomes lie in a more than 2-dimensional simplex was not widely considered by other researchers, so the comparison of theoretical bounds can not be performed. Kivinen and Warmuth's work \citet{Kivinen2001} includes the case when the possible outcomes lie in a more than 2-dimensional simplex and their algorithm competes with all logistic regression functions. They use the relative entropy loss function $\mathcal{L}$ and get a regret term of the order $O(\sqrt{\mathcal{L}_T(\alpha)})$ which is upper unbounded in the worst case. Their prediction algorithm is not computationally efficient and it is not clear how to extend their results for the case when the predictors lie in an RKHS.

We can prove lower bounds for the regret term of the order $O(\frac{d-1}{d} \ln T)$ for the case of the linear model~\eqref{eq:experts1} using methods similar to ones described in \citet{VovkCOS}, and lower bounds for the regret term of the order $O(\sqrt{T})$ for the case of RKHS. Thus we can say that the order of our bounds by time step is optimal. Multiplicative constants may possibly be improved though.

\subsection*{Acknowledgments}\label{sec:acknowledgments}
Authors are grateful for useful comments and discussions to Alexey Chernov, Vladimir Vovk, and Alex Gammerman. This work has been supported by EPSRC grant EP/F002998/1 and ASPIDA grant from the Cyprus Research Promotion Foundation.

\bibliographystyle{plainnat}

\end{document}